\pdfoutput=1

\documentclass[11pt]{article}

\usepackage[]{acl}

\usepackage{times}
\usepackage{latexsym}
\usepackage[T1]{fontenc}

\usepackage[utf8]{inputenc}

\usepackage{microtype}

\usepackage{microtype} 
\usepackage{tabularx}
\usepackage{multirow}
\usepackage{hyperref}
\usepackage{amsmath}
\usepackage{amsfonts}
\usepackage{graphicx}
\usepackage{tikz}
\usepackage{mathtools}
\usepackage{caption}
\usepackage{subcaption}

\newcommand{\bb}{\mathbb}
\newcommand{\cc}{\mathcal}

\title{Benchmarking Neural Network Generalization for Grammar Induction}

\author{
Nur Lan\textsuperscript{1,2}, Emmanuel Chemla\textsuperscript{1}, Roni Katzir\textsuperscript{2}
\\\textsuperscript{1}Ecole Normale Supérieure 
\\\textsuperscript{2}Tel Aviv University
\\ 
\texttt{\{nur.lan,emmanuel.chemla\}@ens.psl.eu}
\\ \texttt{rkatzir@tauex.tau.ac.il}}

\begin{document}

\maketitle

\begin{abstract}
How well do neural networks generalize? Even for grammar induction tasks, where the target generalization is fully known, previous works have left the question open, testing very limited ranges beyond the training set and using different success criteria.
We provide a measure of neural network generalization based on fully specified formal languages. Given a model and a formal grammar, the method assigns a generalization score representing how well a model generalizes to unseen samples in inverse relation to the amount of data it was trained on. The benchmark includes languages such as $a^nb^n$, $a^nb^nc^n$, $a^nb^mc^{n+m}$, and Dyck-1 and 2. We evaluate selected architectures using the benchmark and find that networks trained with a Minimum Description Length objective (MDL) generalize better and using less data than networks trained using standard loss functions. The benchmark is available at \href{https://github.com/taucompling/bliss}{https://github.com/taucompling/bliss}. 

\end{abstract}

\section{Introduction}
\label{sec:intro}

\begin{table*}[ht!]
\begin{tabularx}{\textwidth}{XXXllll}

\hline
\multicolumn{1}{l}{\textbf{Language}} & \multicolumn{1}{l}{\textbf{Paper}} & \multicolumn{1}{l}{\textbf{Model}} & \multicolumn{1}{l}{\textbf{Metric}} & \multicolumn{1}{l}{\textbf{Training size}} & \multicolumn{1}{l}{\textbf{Max train $n$}} & \multicolumn{1}{l}{\textbf{Max test $n$}} \\
\hline
\multirow{4}{*}{$a^nb^n$} & GS'01 & LSTM & $M_{cat'}$ & $16{,}000$ & $30$ & $1{,}000$ \\ \cline{2-7}
 & JM'15 & Stack-RNN & $M_{det}$ & $20^\dagger$ & $19$ & $60$ \\ \cline{2-7} 
 & WGY'18 & LSTM & $Bin$ & $100^\dagger$ & $100$ & $256$ \\ \cline{2-7} 
 & LGCK'22 & MDLRNN & $M_{det}$ & $500$ & $22$ & $\infty$ \\ \hline 
\multirow{4}{*}{$a^nb^nc^n$} & GS'01 & LSTM & $M_{cat'}$ & $51{,}000$ & $40$ & $500$ \\ \cline{2-7} 
 & JM'15 & Stack-RNN & $M_{det}$ & $20^\dagger$ & $19$ & $60$ \\ \cline{2-7} 
 & WGY'18 & LSTM & $Bin$ & $50^\dagger$ & $50$ & $100$ \\ \cline{2-7} 
 & LGCK'22 & MDLRNN &  $M_{det}$ & $500$ & $22$ & $\infty$ \\ \hline
\multirow{4}{*}{Dyck-1} & SGBS'19a & LSTM & $M_{cat'}$ & $10{,}000$ & $50$ & $100$ \\ \cline{2-7} 
 & SGBS'19b & MARNN & $M_{cat'}$ & $5{,}000$ & $50$ & $100$ \\ \cline{2-7}
    & EMW'22 & ReLU-RNN & $M_{cat'}$ & $10{,}000$ & $50$ & $1{,}000$ \\ \cline{2-7}

 & LGCK'22 & MDLRNN & $M_{cat}$ & $500$ & $16$ & $\infty$ \\ \hline
\end{tabularx}

\caption{ANN performance in selected probes of formal language learning. \textbf{Metrics} (see Section~\ref{sec:accuracy-metrics}): $M_{det}$ = deterministic accuracy; $M_{cat}$ = categorical accuracy; $M_{cat'}$ = a non-probabilistic version of $M_{cat}$; $Bin$ = binary classification from hidden state to accept/reject labels, based on positive and negative samples.
\textbf{Training size}: $\dagger$ = the paper did not explicitly specify the training set size, the value here is derived by assuming the training set was an exhaustive list of all strings up to `max train $n$`.
\textbf{`Max test $n$'}: the largest $n$ for which the criterion was reached. For Dyck-1, $n$ represents overall sequence length. `$\infty$' = the paper provides evidence that the network is correct for any $n$. 
When a paper reports several experiments as in GS'01, we take the best result based on the smallest training set.
\textbf{Papers}: GS'01 = \citet{gers:2001}; JM'15 = \citet{joulin:2015}; WGY'18 = \citet{weiss:2018}; SGBS'19a = \citet{suzgun:2019a}; SGBS'19b = \citet{suzgun:2019}; EMW'22 = \citet{el-naggar:2022}; LGCK'22 = \citet{lan:2022a}.}
\label{table:probes}
\end{table*}

The extent to which artificial neural networks (ANNs) generalize beyond their training data is an open research question.
In this work we approach this question from the perspective of grammar induction, that is, the learning of a formal grammar from a finite (often small) sample from the (typically infinite) language of that grammar. In order to succeed in this task, a model must strike a balance between fitting the training data and generalizing to a potentially infinite set of unseen strings.
Humans tested on such tasks show systematic generalization from small sets of examples (\citealp{fitch:2004}, \citealp{malassis:2020}).

While a range of ANN architectures have been shown to reach approximations for formal languages, the quality of this approximation remains an open matter, as we show below. 
Here we build on previous probes of ANN generalization for grammar induction and introduce a unified and general way to assess this capability, for a given pair of a learning model and a corpus drawn from a formal language.
Our main contributions are:
    
    \begin{enumerate}
        
        \item \textbf{A benchmark for formal language learning}. The benchmark relies on a method for quantifying ANN generalization for formal languages, including probabilistic languages.
        The method assigns an index score representing a model's generalization performance in inverse relation to the size of the training data. We introduce the method and provide concrete datasets for well-studied formal languages.
        
        \item \textbf{An evaluation of selected architectures.} We test recurrent neural networks (RNNs) of the Long-Short Term Memory  type (LSTM; \citealp{hochreiter:1997}); Memory-augmented RNNs (MARNN; \citealp {suzgun:2019};) and an RNN variant which replaces the traditional gradient-based training regime with an objective that optimizes the model's Minimum Description Length (MDLRNN; \citealp{lan:2022a}).

        We find that equipping ANNs with memory devices such as differentiable stacks helps generalization, but generalization remains partial and slow. At the same time, training with MDL leads in some of the test cases that we examined to potentially perfect generalization with significantly less data. In other cases, training with MDL did not result in successful generalization, possibly because the optimization procedure we used for the architecture search failed to find the global optimum under the MDL objective function.

    \end{enumerate}

\section{Background}

Learning formal languages has long been used to probe various aspects of ANN performance. These most often include inquiries about: (i) ANNs' ability to generalize beyond their training data, and (ii) ANNs' expressive power; i.e., whether they can represent the relevant target grammars (often probed with reference to the Chomsky hierarchy of formal languages, as in \citealp{deletang:2022}). Here we will focus on the generalization question. We will show how it might be related to another under-exploited line of inquiry regarding the training objective of ANNs.

A long line of theoretical work has probed the computational power of ANNs. \citet{siegelmann:1992} originally showed that RNNs with a sigmoid activation can emulate multiple-stack Turing machines under certain permissive conditions (infinite activation precision and unbounded running time). Since these conditions cannot be met in practice, another line of work probed the computational power of RNNs under practical conditions (finite precision and input-bound running time). \citet{weiss:2018} have shown that under these conditions LSTMs are able to hold weight configurations that perform unbounded counting, and so they should be able to recognize counter languages (CL), a family of formal languages that can be recognized using one or more counting devices (following some formal restrictions, \citealp{merrill:2021}). Recently, \citet{el-naggar:2023} and \citet{el-naggar:2023a} have shown that two simpler RNN architectures, with linear- and ReLU-based cells, are also able to hold counting weight configurations, with similar consequences for recognizing CL.

Empirically, another line of work provided promising results regarding the capability of ANNs to learn formal languages. This was most often done by training networks on strings up to a certain length and then showing good performance on longer ones (\citealp{boden:2000}, \citealp{gers:2001}; see Table~\ref{table:probes}). \citet{gers:2001} have shown that LSTMs trained on languages such as $a^nb^n$ and $a^nb^nc^n$ with $n$ values in the low dozens perform well on $n$'s in the high hundreds. \citet{suzgun:2019a} found that LSTMs trained on Dyck-1 sequences (strings of well-balanced pairs of brackets) up to length 50 performed well on lengths up to 100. \citet{suzgun:2019} proposed RNN variants that are equipped with external differentiable memory devices and showed that they yield improved performance on non-regular languages.

However, other empirical results show that in practice ANNs generalize only to very restricted ranges. \citet{weiss:2018} found that while LSTMs are theoretically able to hold counting solutions, these are not found through training: LSTMs trained on $a^nb^n$ and $a^nb^nc^n$ with max $n$ 100 and 50, respectively, start accepting illicit strings with $n$ values as low as 256 and 100. 
As mentioned above, \citet{suzgun:2019a} tested LSTMs on Dyck-1 sequences but only up to length 100, and concluded that this language was learned by LSTMs. \citet{el-naggar:2022} extended this work to longer sequences, and found that LSTMs fail to generalize in practice, outputting incorrect predictions at lengths under 1,000. This, despite Dyck-1 being a CL and so theoretically learnable by LSTMs \cite{weiss:2018}.

Apart from LSTMs, recent probes by \citet{el-naggar:2023} and \citet{el-naggar:2023a} have shown that linear and ReLU RNNs, theoretically capable of counting, fail to find the counting weight configurations in practice when trained using backpropagation and standard loss functions; \citet{el-naggar:2023a} went further with determining the source of this discrepancy, showing that the counting weight configuration is not an optimum of these loss functions.

Moreover, even in works that report successful generalization to some degree beyond the training set, the fact that networks stop generalizing at an arbitrary point is often left unexplained 
(\citealp{gers:2001}, \citealp{suzgun:2019a}, \citeyear{suzgun:2019}, \citealp{deletang:2022}, a.o.).\footnote{Technical limitations such as finite activation precision can be ruled out as explanations for generalization failures, at least for counter languages and models where network states serve as memory: as shown in works mentioned above, ANNs often start outputting wrong predictions for $n$ values in the low hundreds. Even restricted representations such as 16-bit floats can hold much larger values, and modern implementations such as PyTorch use 32-bit floats by default.}

The literature on the generalization abilities of ANNs has made use of a range of measures of success, making results difficult to compare. Different probes of the same model often use different success criteria, and generate training and test sets using different sampling methods and of different orders of magnitude.
Table~\ref{table:probes} summarizes selected probes of ANN generalization and highlights the fragmented nature of this literature.
In the following sections we propose a unified method to consolidate these efforts and better understand the generalization capabilities of ANNs.

\section{The \textsc{bliss} index}

We present the Benchmark for Language Induction from Small Sets (\textsc{bliss}). We provide a formal description of the method, followed by a concrete application to specific tasks.
        
The current release consists of three parts: (i) A specification for the generalization index $\bb{B}$, calculated for a given pair of formal language and ANN; (ii) A dataset containing a set of formal languages for benchmarking; (iii) An evaluation of different ANN architectures using this dataset.

\subsection{General setting: models and tasks}
\label{sec:index-definition}

For a given model $A$, e.g., an LSTM, a task is composed of the following components:

\begin{itemize}
\setlength\itemsep{0em}
\item $G$ -- a grammar, e.g., a probabilistic context-free grammar (PCFG).

\item $S$ -- a sampling method from $\cc{L}(G)$, the language generated by $G$.

\item $\cc{C} = S(G)$ -- a training corpus, may contain repetitions. 

\item $\cc{T} \subseteq \cc{L}(G) \setminus \cc{C}$ -- a test corpus.

\item $M$ -- a task-specific accuracy metric with adjustable error margin 

$\varepsilon\in[0,1]$. 
It uses predictions $A(s)$ on strings $s \in \cc{T}$ to calculate an accuracy score $M(A,\cc{T}, \varepsilon) \in [0,1]$.
	
	\item $N$ -- a task-specific constant for setting the order of magnitude of dataset sizes. For example, $N=3$ sets the order of magnitude at $10^3$. Training and test sizes are then derived as described below. Selecting $N$ is done empirically based on properties of the task, e.g., languages with large vocabularies require larger amounts of training data, hence a larger $N$. 

\end{itemize}		
	
\subsection{From task to generalization index}

\begin{figure}[!t]
\centering
\includegraphics[scale=0.53]{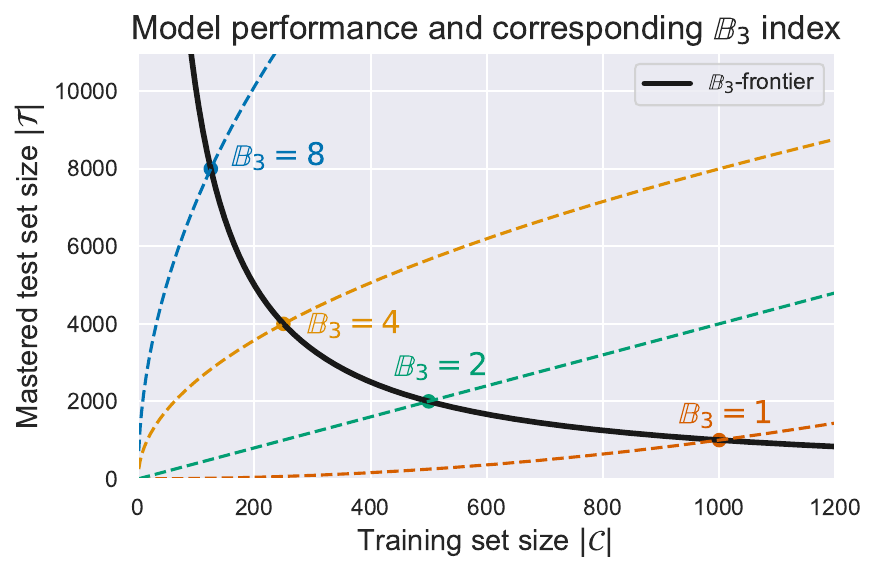}
\caption{Example generalization index scores $\bb{B}_3$, i.e., for a baseline training size of $10^3$. Each dashed line represents the performance profile of some hypothetical model, as a function of the size of the training set. The intersection with the $\bb{B}_3$-frontier indicates its $\bb{B}_3$ index.
}

\label{fig:b-generalization-index}
\end{figure}

For a given task, the generalization index of order $N$ for a model $A$ is then defined as:
	\begin{equation}			
	\label{eq:index-def}
	\bb{B}_{N}^{\cc{L}}(A) = max \Biggl\{ b\ \bigg| \	
		\begin{array}{c}
			|\cc{T}| = 10^N \times b, \\
			|\cc{C}| = 10^N /~b, \\			
			M(A, \cc{T}, \varepsilon) = 1.0
		\end{array}	
	 \Biggr\}
		\end{equation}

Intuitively, the index compares a model's performance on a test size |$\cc{T}$| to the inverse of its training data size |$\cc{C}$|. 

The index is expressed as the maximal $b$ factor
which scales the training set and the corresponding test set in opposite directions: The accuracy condition at the bottom of (\ref{eq:index-def}) means that the model should be $\varepsilon$-close to perfect generalization on the test set. A model's generalization index $\bb{B}$ thus represents the performance that can be maximally `squeezed out' of an inversely small amount of data. 

Figure~\ref{fig:b-generalization-index} exemplifies selected $\bb{B}$ values calculated based on~(\ref{eq:index-def}).
For illustration, for $a^nb^n$, using the order of magnitude $N=3$, a model that was trained on $|\cc{C}| = 10^3/2 = 500$ samples and was 100\% accurate on a test set of size $10^3\times 2=2,000$ will have an index score $\bb{B}_{3}^{a^nb^n} \ge 2$. A  model for the same language that was trained on 250 samples only and generalized to a subsequent set of 4,000 samples will reach $\bb{B}_{3}^{a^nb^n} \ge 4$.
		
For practical reasons, one cannot exhaust all values of $b$ to find $\bb{B}$. However, training and evaluating a model using a few $b$ values is enough to reveal its generalization dynamics, as shown in experiments in Section~\ref{sec:experiments}. The following sections describe the specific choices made for the different benchmark components in these experiments.

\subsection{Learning setup}
\label{sec:language-modeling}

Previous work surveyed here differed in their learning setup. \citet{gers:2001} and \citeauthor{suzgun:2019a} (\citeyear{suzgun:2019a}, \citeyear{suzgun:2019}) trained networks in a non-probabilistic, supervised setup by exposing the model to all possible next symbols and minimizing the mean-squared error (MSE) -- i.e., the model is given explicit information about the distribution of possible symbols. \citet{joulin:2015} and \citet{lan:2022a} used a setup that we adopt below, in which model outputs are probabilistic, and training is self-supervised language modeling (i.e., the model is exposed to the next symbol only) with a cross-entropy loss. \citet{weiss:2018} trained a binary classifier with accept/reject labels based on positive and negative examples.

Since our focus is grammar induction, here we adopt the more demanding setup of learning from positive examples alone. All tasks are thus designed as self-supervised language modeling. At each time step, a model assigns a probability distribution to the next symbols in the string.

The benchmark is agnostic as to the internals of the model and its training, as long as its outputs represent a probability distribution over symbols. In practice, then, the method can be applied to any language model, not necessarily an ANN.

\subsection{Sampling}
\label{sec:sampling-method}
	
To construct the training and test sets $\cc{C}$ and $\cc{T}$ we use the following as method $S$:
	
	\begin{itemize}
		
	\item To construct $\cc{C}$, we sample strings according to the distribution defined by $G$, with repetitions. For example, if $G$ is a PCFG, it can be sampled by applying derivation rules chosen proportionally to their expansion probabilities. Repetitions are allowed so that $\cc{C}$ follows a similar surface distribution to $\cc{L}(G)$ and so that the model can pick up on the underlying probabilities in $G$. 
	
	\item To construct $\cc{T}$, we take the $|\cc{T}|$ subsequent strings starting right after the longest string in $\cc{C}$, sorted by length.\footnote{Test strings may need to be sorted further according to specific properties of a language, see Section~\ref{sec:training_test_sets}.} For example, for the language $a^nb^n$, if the longest string in the training set $\cc{C}$ was $a^{17}b^{17}$, and the model needs to be tested on a set of 2000 strings, $\cc{T}$ will be composed of the strings $a^{18}b^{18}, ..., a^{2017}b^{2017}$.	
	
\end{itemize}	

The sampling method $S$ can be either probabilistic as described here, or exhaustive, training on all strings in $\cc{L}$ up to a certain length. We opt for probabilistic sampling because of the nature of the task at hand: the models under discussion here are trained to assign probabilities to the next symbol in a string, most often minimizing a cross-entropy loss. In practice, then, they always learn distributions over strings. Thus if $\cc{C}$ follows a similar surface distribution to $\cc{L}$ (given a large enough sample size), the model should eventually learn this distribution in order to minimize its loss.

Probabilistic sampling thus makes it possible to probe both a model's knowledge about the surface forms of $\cc{L}$ (by treating model outputs as categorical classes), and about their distribution. The modularity of the index makes it possible to choose either option by varying the accuracy metric $M$, as we show in the next section.

\begin{figure}[b!]
  \centering
  \[
  \begin{array}{lccccccc}
    \hline \\
  
\text{Input:} & \# &   $($ &   $($ &   $)$  &  $($ & $)$ & $)$ \\
  & \downarrow &   \downarrow &   \downarrow &   \downarrow &   \downarrow &   \downarrow &   \downarrow \\
    \text{Target:} & \#/$($ &   $($/$)$ &  $($/$)$ &  $($/$)$ & $($/$)$ & $($/$)$ & \#/$($ \\ \\
        \cline{1-8}       
\\
  \text{Input:} & \# &   a &   a &   a  &  b & b & b \\
  & \downarrow &   \downarrow &   \downarrow &   \downarrow &   \downarrow &   \downarrow &   \downarrow \\
    \text{Target:} & \#/a &   a/b &   a/b &   a/b & \pmb{b} & \pmb{b} & \pmb{\#} \\
    & & & & & \multicolumn{3}{c}{\underbrace{\makebox[1.3cm]{}}_{\text{Deterministic}}} \\
    \cline{1-5}   
  \end{array}
  \]
  \caption{Inputs and valid next symbols at each step of a Dyck-1 string (top) and $a^nb^n$ (bottom), including the start/end-of-sequence symbol `$\#$'. For $a^nb^n$, accuracy is measured at deterministic steps, after the first `b'. For Dyck-1, accuracy is  the fraction of time steps where a model predicts only valid next symbols: `$\#$' should be predicted only when brackets are well balanced.}
  
  \label{fig:example-strings}
\end{figure}

\subsection{Accuracy metrics}
\label{sec:accuracy-metrics}

Ultimately we are interested in knowing whether a model accepts all strings in $\cc{L}$ and rejects all others. In classical formal language theory, where discrete automata are used, acceptance is clear cut and taken as going into an accepting state. ANNs on the other hand use continuous representations with no standard acceptance criterion.

Different acceptance criteria have been used in previous works to measure success for ANNs: \citet{gers:2001} and \citet{suzgun:2019} defined acceptance of a string as a model assigning output values above a certain threshold to valid symbols only; \citet{joulin:2015} measure accuracy at parts of strings that are completely predictable;
and \citet{weiss:2018} turn a network into a recognizer by training a binary classifier from network states to accept/reject labels. Below we provide general versions of these accuracy metrics (omitting \citealp{weiss:2018} who rely on negative examples).

Choosing which metric to use is based on the properties of the language at hand. Well-performing models might still deviate slightly from perfect accuracy due to practical limitations, such as a softmax function preventing a model from expressing categorical decisions. Thus for each accuracy metric we add an adjustable error margin $\varepsilon$. Acceptance of a string is defined as reaching 100\% accuracy (minus $\varepsilon$) on the string. Success on the test set is then defined as accepting all strings in the set (third condition in (\ref{eq:index-def})).

	\begin{enumerate}
		\item \textit{Deterministic accuracy} ($M_{det}$). Some languages contain strings with deterministic phases, where the next symbol is fully predictable. For example, strings in the language $a^nb^n$ have two phases, the $a$ phase and the $b$ phase. As long as only $a$'s are seen, the next symbol remains unpredictable as the sequence can continue with another $a$ or switch to the $b$ phase. The string becomes deterministic once the first $b$ appears. $M_{det}$ is defined as the fraction of deterministic time steps in which the model assigns the majority probability to the correct next symbol. This metric is used in \citet{joulin:2015}.
		
	A string is considered accepted if the model is $1-\varepsilon$ accurate over all deterministic time steps. Note however that even a very small $\varepsilon$ might benefit models that do not recognize strings well. For example, for the language $a^nb^n$, the deterministic steps in a string are the $b$'s and the final end-of-sequence symbol. A degenerate model that predicts only $b$'s will get only the end-of-sequence symbol wrong out of all deterministic steps, and will reach a very high accuracy score. For any large enough test set these errors will be hidden within the $\varepsilon$ margin and the model will be deemed successful. $\varepsilon$ should therefore be chosen with care per task.

 $M_{det}$ is used below for the following languages that have deterministic phases: $a^nb^n$, $a^nb^nc^n$, $a^nb^nc^nd^n$, and $a^nb^mc^{n+m}$.
		
		\item \textit{Categorical accuracy} ($M_{cat}$). Some language strings do not have any predictable phases. This is the case in the Dyck family of languages. At each time step in a Dyck string, one may open a new bracket (see Figure~\ref{fig:example-strings}). $M_{cat}$ is therefore defined as the fraction of steps in which a network assigns probability $p > \varepsilon$ to each possible next symbol, and $p \leq \varepsilon $ to irrelevant symbols. Non-probabilistic versions of $M_{cat}$ are used in \citet{gers:2001} and \citeauthor{suzgun:2019} (\citeyear{suzgun:2019a}, \citeyear{suzgun:2019}) who do not treat network outputs as probability distributions. $M_{cat}$ is used below for Dyck languages.
  
\end{enumerate}
			
As specified in Section~\ref{sec:index-definition}, the index $\bb{B}$ is calculated based on the largest test set for which a model reaches an $\varepsilon$-perfect accuracy score. 

Beyond accuracy, one might be interested in inspecting a model's knowledge of the distribution of strings in $\cc{L}$ induced by a probabilistic $G$. This can be done by using the probabilistic sampling method described in Section~\ref{sec:sampling-method} and accompanying it with a probabilistic accuracy measure -- for example, one based on an optimal cross-entropy score, which is known from $G$'s expansion probabilities (as done in \citealp{lan:2022a}). Feeding loss values into an accuracy metric will require normalizing them across tasks. We leave this extension for future work.

\subsection{String structure}	 
	    
Following \citet{gers:2001}, each sequence starts and ends with a start/end-of-sequence symbol `$\#$'. This turns the task into a strict acceptance/rejection task -- predicting the end-of-sequence symbol is taken as going into an accept state.
The start- and end-of-sequence symbols are added to the task-specific vocabulary and are assigned probabilities by the model at each step. Figure~\ref{fig:example-strings} illustrates input and target sequences for $a^nb^n$ and Dyck-1.

\subsection{Limitations}
	
One shortcoming of the proposed index score is that it does not reflect perfect generalization, i.e., it is an empirical index that cannot point out a model that outputs correct predictions for \textit{any} string in $\cc{L}(G)$. For most models, this will not be a problem, and $\bb{B}$ will simply represent the model's best training vs.~test size ratio. In the case of a model that reaches perfect generalization on any input, the index score will represent the critical training size that brings the model to this performance.

	Assigning a generalization score to infinitely correct models will remain a problem for any empirical metric that assigns scores to models based on finite test values. An alternative to such empirical probes would be to analytically show that a model is correct (as done in \citealp{lan:2022a}).

\section{Datasets}
\label{sec:datasets}

We provide training and test datasets for a preliminary set of formal languages for evaluation using the $\bb{B}$ index. The dataset includes the languages $a^nb^n$, $a^nb^nc^n$, $a^nb^nc^nd^n$, $a^nb^mc^{n+m}$, Dyck-1, and Dyck-2. 
The source code, datasets, and specifications for the benchmark are available at \href{https://github.com/taucompling/bliss}{https://github.com/taucompling/bliss}.

\subsection{Training and test sets}
\label{sec:training_test_sets}

Training sets for context-free languages are sampled from PCFGs as described in Section~\ref{sec:sampling-method}. The PCFGs are given in Appendix~\ref{appendix:pcfgs}. Training sets for context-sensitive languages are generated by sampling values for $n$ from a geometric distribution. 

Test sets are generated using the method described in Section~\ref{sec:sampling-method}: All test sets consist of an exhaustive list of strings ordered by length starting right after the longest string seen during training. Test sets for $a^nb^mc^{n+m}$ consist of the list of strings starting after the last seen pair of $n,m$, sorted by $n+m$ values to test all possible combinations. 	

\section{Experiments}
\label{sec:experiments}

\subsection{Models}
\label{sec:experiments:models}

We test the following models: LSTM RNNs \cite{hochreiter:1997}; Memory-augmented RNNs (MARNN; \citealp{suzgun:2019a}); and Minimum Description Length RNNs (MDLRNN; \citealp{lan:2022a}).

LSTM architectures were developed with the task of keeping items in memory over long distances in mind. As mentioned above, \citet{weiss:2018} have shown that LSTMs are theoretically capable of recognizing CL. 

MARNNs \cite{suzgun:2019} are RNNs equipped with external memory devices, and were shown to yield better performance when learning languages that require stack-like devices and beyond. Here we use Stack-LSTM, an LSTM augmented with a pushdown automaton; and Baby Neural Turing Machines (Baby-NTM; itself a variant of NTMs, \citealp{graves:2014}), an RNN with a more freely manipulable memory.\footnote{We modify \citet{suzgun:2019a}'s models to output probability distributions, replacing the final sigmoids with a softmax layer and the MSE loss with cross-entropy. See Section~\ref{sec:language-modeling}.}

MDLRNNs are RNNs trained to optimize the Minimum Description Length objective (MDL; \citealp{rissanen:1978}), a computable approximation of Kolmogorov complexity, the algorithmic complexity of a model. The intuition behind the objective is equating compression with finding regularities in the data: a model that compresses the data well will generalize better and avoid overfitting. In practice, optimization is done by minimizing the sum of the architecture encoding length and the standard cross-entropy loss, both measured in bits based on a specific encoding scheme.

MDL is a stricter regularizer than standard regularization techniques such as L1/L2: the latter penalize large weight values but cannot prevent models from overfitting using a solution that uses many small, but informative, weights. MDL penalizes the actual information content of the network, forcing it to be general and avoid overfitting. MDLRNNs were shown to learn some of the languages discussed here in full generality using small architectures of only 1 or 2 hidden units and to outperform L1/L2 \cite{lan:2022a}.

MDL is a non-differentiable objective, which requires that MDLRNN be optimized using a non-gradient based search method, such as an evolutionary algorithm that searches the network architecture space.
Since this method is not based on gradient descent, \citet{lan:2022a} were able to use non-standard, non-differentiable activations such as step functions. Here we restrict the architecture space to only standard activations: the linear function, ReLU, and tanh. This serves both to compare MDLRNN with standard networks and to limit the architecture search space. We publish the resulting nets as part of the MDLRNN-Torch release at \href{https://github.com/0xnurl/mdlrnn-torch}{https://github.com/0xnurl/mdlrnn-torch}.

Appendix~\ref{appendix:hyperparams} lists the hyper-params for all runs.    

\subsection{Training sets}
\label{sec:training-sets}
We used training sizes $|\cc{C}| = 100, 250, 500, 1000$. 
We stopped at the smallest size 100 because in our setup this size results in test strings of lengths $>10{,}000$, leading to very long running times.

\subsection{Index parameters}

We calculate the $\bb{B}$ index for all trained networks using the following index parameters:

Magnitude parameter $N=3$, i.e., training and test sizes are derived from a baseline size $10^3$. This order of magnitude was selected based on the training set sizes used in previous works for the languages inspected here (Table~\ref{table:probes}).

$M_{det}$ $\varepsilon=0.005$, i.e., a model needs to correctly predict the next symbol for at least 99.5\% of all deterministic steps. Since even this high threshold allows a degenerate model to reach good scores as described in Section~\ref{sec:accuracy-metrics}, we also calculate the index score using $\varepsilon=0$, i.e.\ a model must predict \textit{all} deterministic symbols correctly.

$M_{cat}\ \varepsilon=0.005$, i.e., for Dyck, a model needs to assign  $p\le0.005$ to each irrelevant symbol and $p > 0.005$ to possible ones. Here as well we report results for $\varepsilon=0$, i.e., a model must assign non-zero probabilities to valid symbols only.

\section{Results}

\subsection{Non-perfect accuracy}

\begin{table}[t]
\begin{tabularx}{\columnwidth}{XXcc}
\hline
\multirow{2}{*}{\textbf{Language}} & \multirow{2}{*}{\textbf{Model}} & \multicolumn{2}{c}{\large $\bb{B}$-index} \\
 &  & \multicolumn{1}{l}{\footnotesize $\varepsilon=0.005$} & \multicolumn{1}{l}{\footnotesize $\varepsilon=0$} \\ \hline

\multirow{3}{*}{$a^nb^n$} & LSTM & \textbf{10} & <1 \\ \cline{2-4}
 & \textls[-15]{Stack-LSTM} & \textbf{10} & <1 \\ \cline{2-4}
 & Baby-NTM & \textbf{10} & 1 \\ \cline{2-4} 
 & MDLRNN & \textbf{10} & \textbf{10}   \\ \hline

\multirow{3}{*}{$a^nb^nc^n$} & LSTM & <1 & <1 \\ \cline{2-4} 
 & \textls[-15]{Stack-LSTM} & 2 & <1 \\ \cline{2-4}
 & Baby-NTM & \textbf{10} & <1 \\ \cline{2-4} 
 & MDLRNN & <1 & <1 \\ \hline
 
\multirow{3}{*}{$a^nb^nc^nd^n$} & LSTM & <1 & <1 \\ \cline{2-4} 
 & \textls[-15]{Stack-LSTM} & 1 & <1 \\ \cline{2-4}
 & Baby-NTM & \textbf{4} & <1 \\ \cline{2-4} 
 & MDLRNN & <1 & <1 \\ \hline
 
\multirow{3}{*}{$a^nb^mc^{n+m}$} & LSTM & <1 & <1 \\ \cline{2-4} 
 & \textls[-15]{Stack-LSTM} & \textbf{10} & <1 \\ \cline{2-4}
 & Baby-NTM & 4 & <1 \\ \cline{2-4} 
 & MDLRNN & 4 & \textbf{4} \\ \hline
 
\multirow{3}{*}{Dyck-1} & LSTM & <1 & <1 \\ \cline{2-4} 
 & \textls[-15]{Stack-LSTM} & <1 & <1 \\ \cline{2-4}
 & Baby-NTM & <1 & <1 \\ \cline{2-4} 
 & MDLRNN & \textbf{2} & \textbf{2} \\ \hline
 
\multirow{3}{*}{Dyck-2} & LSTM & <1 & <1 \\ \cline{2-4} 
 & \textls[-15]{Stack-LSTM} & \textless{}1 &  <1 \\ \cline{2-4}
 & Baby-NTM & \textless{}1 &  <1 \\ \cline{2-4} 
 & MDLRNN & \textless{}1 &  <1 \\ \hline
\end{tabularx}

\caption{Generalization scores $\bb{B}$. The index represents how well a model generalizes in relation to its training size. A score $\bb{B} = 4$ indicates that a model trained on 250 samples reached the accuracy criterion on the consecutive 4,000 unseen test samples. $\bb{B} < 1$ indicates that the model did not reach the accuracy criterion when the test size was greater than the training size, but might reach it for larger training and smaller test sets.}

\label{table:index-accuracy}

\end{table}

The generalization index obtained by each model for each language is presented in Table~\ref{table:index-accuracy}. 
	
We start by inspecting the indexes calculated using the more lenient accuracy margin $\varepsilon=0.005$. 

For $a^nb^n$, under this accuracy margin, all models are assigned index $\bb{B}=10$, i.e., reaching the success criterion for the next unseen 10,000 samples after being trained on 100 samples. For the specific combination of random seed and sampling prior in these experiments, this means that the models were trained on strings up to $a^{20}b^{20}$ and generalized to all strings up to at least $a^{10020}b^{10020}$ with deterministic accuracy $M_{det} \ge 99.5\%$.

For $a^nb^nc^n$, MARNNs reach $\bb{B}=$ 10 and 2, while LSTM and MDLRNNs do not reach the success criterion, resulting in $\bb{B}<1$. For $a^nb^nc^nd^n$ only MARNNs reach a specified index, with a Baby-NTM reaching $\bb{B}=4$, indicating that it generalized to strings as long as $a^{4020}b^{4020}c^{4020}d^{4020}$ with $M_{det} \ge 99.5\%$.

For the addition language $a^nb^mc^{n+m}$, Stack-LSTM and MDLRNN reached index scores $\bb{B}=$ 10 and 4 respectively. For the specific combination of random seed and the sampling prior used here, this means that the winning Stack-LSTM saw maximum values of $n=18, m=20$ during training, and generalized to all strings up to $a^{120}b^{120}c^{240}$ with $M_{det} \ge 99.5\%$.

\subsection{Perfect accuracy}

\begin{figure}[]
\centering
\includegraphics[scale=0.51]{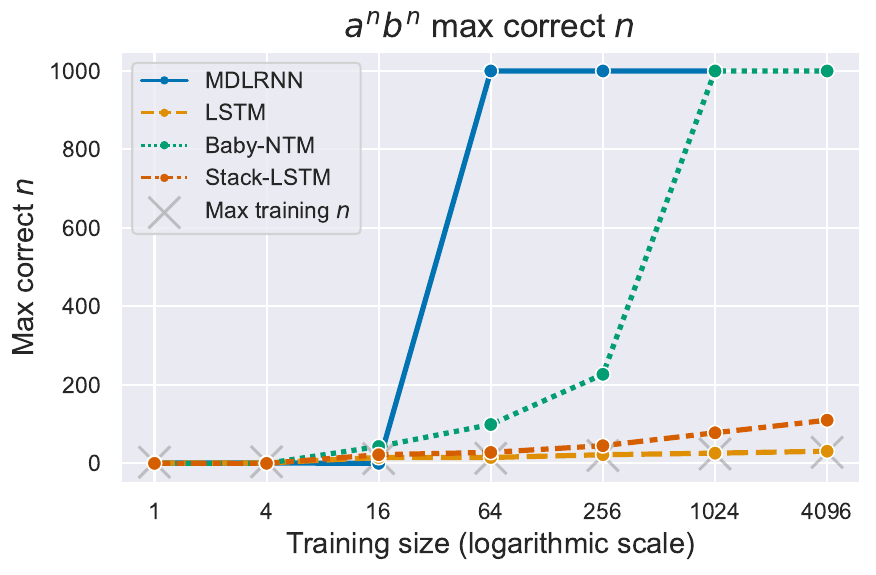}
\caption{Generalization performance of the models tested here. Models were trained on strings drawn from $a^nb^n$ and tested on acceptance of strings up to $n=1{,}000$. \textit{X}'s mark the maximum $n$ seen during training.}

\label{fig:anbn-max-n}
\end{figure}

We report the generalization scores using a strict $\varepsilon=0$ as well, i.e., when a model is required to predict \textit{all} deterministic steps correctly or assign non-zero probability to valid symbols only. For languages with deterministic steps such as $a^nb^n$, this means that the model needs to always predict the end-of-sequence symbol correctly, thus making a distinction between accepting a string and approximating its surface structure.

Here, only MDLRNNs remain at the same scores, indicating that they predicted all time steps correctly. Baby-NTM reaches $\bb{B}=1$ for $a^nb^n$, a drop from 10. The rest of the networks drop to $\bb{B}<1$, revealing that their good scores in the previous comparison calculated with a non-zero $\varepsilon$ was due to them approximating the target languages, even at low $n$ values.

\begin{figure}[t]
\centering
\includegraphics[scale=0.185]{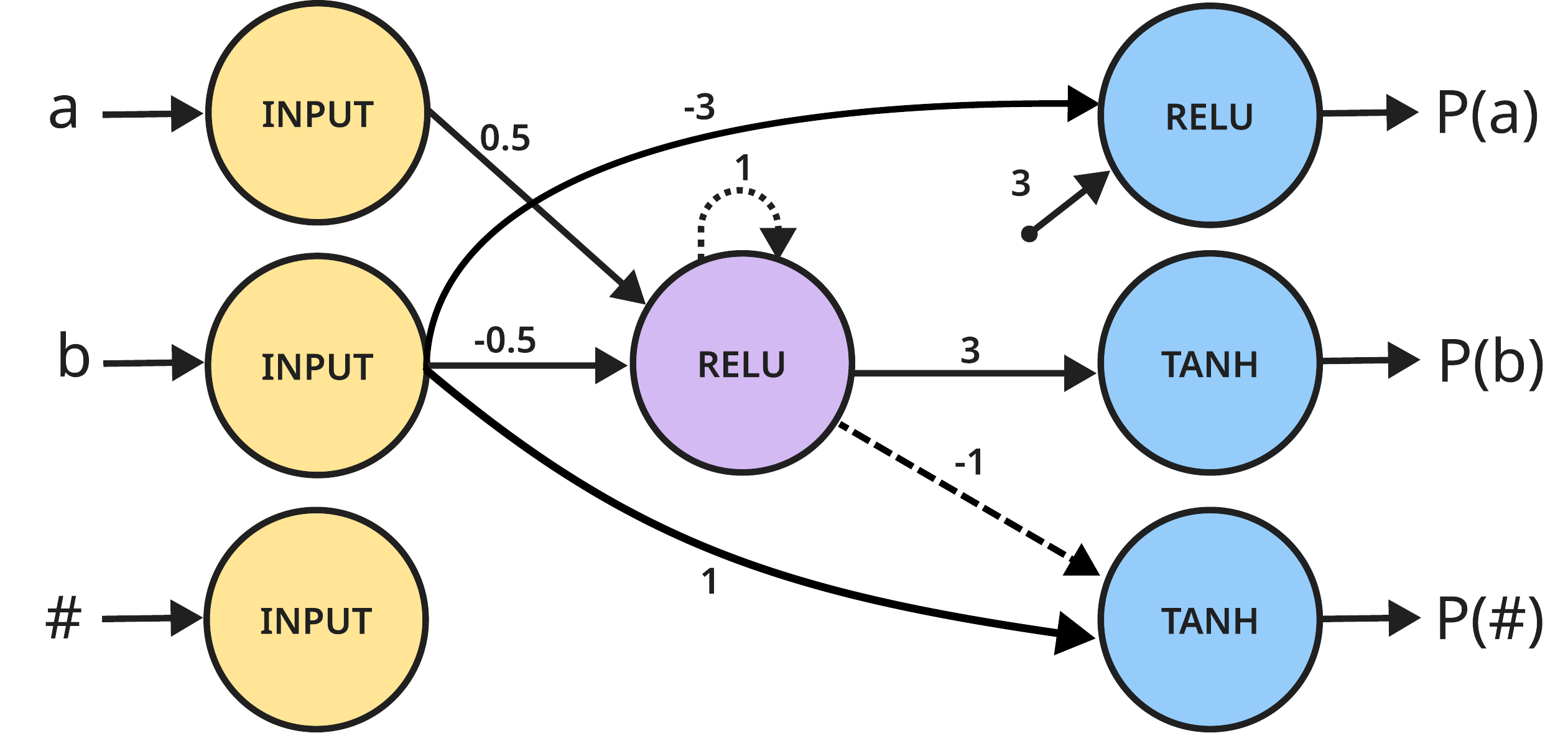}
\caption{RNN cell architecture of the best-performing MDLRNN for $a^nb^n$, which trained on 100 samples and reached $\bb{B}_3=10$. The network uses only one hidden unit and standard activation functions, and generalizes up to at least $a^{35000}b^{35000}$. Dashed arrows are recurrent connections across time steps. The loop from the hidden ReLU unit to itself is a counter mechanism evolved by the evolutionary algorithm to count and compare the number of $a$'s and $b$'s.}

\label{fig:mdlrnn-anbn}
\end{figure}

MDLRNN performance here is in line with results from \citet{lan:2022a}, who provided evidence that MDLRNNs for these languages do not only perform empirically well on large test values, but are also provably correct for any input. However, here we limited activations to standard, non-discrete functions (Section~\ref{sec:experiments:models}), potentially limiting the network's ability to generalize well in the limit. While we do not provide correctness proofs for the networks found here, the index scores indicate that MDLRNNs generalize well to large values using only standard activations. Figure~\ref{fig:mdlrnn-anbn} presents the MDLRNN found for $a^nb^n$. We checked whether this network also accepts $n$ values beyond those needed to reach the score $\bb{B}=10$ ($n=10{,}020$). The network reached 100\% $M_{det}$ for all values up to $n=35{,}000$, at which point we stopped the test due to long feeding times.

Beyond the benchmark scores, Figure~\ref{fig:anbn-max-n} plots the largest $n$ value for $a^nb^n$ strings predicted by the models tested here with 100\% $M_{det}$ accuracy ($\varepsilon=0$), as a function of training set size. Both MDLRNNs and Baby-NTMs reach perfect accuracy up to the tested maximum of $n=1{,}000$. MDLRNNs however require two orders of magnitude less data to reach this performance (and the benchmark scores in Table~\ref{table:index-accuracy} show that in fact MDLRNNs generalized up to at least $n=10{,}000$, while Baby-NTMs remained at $1{,}000$). LSTMs and Stack-RNNs did not generalize well beyond the training samples. 
This is in line with previous works showing that these models may need substantially more training data in order to learn these languages (Table~\ref{table:probes}).

\section{Discussion}

We provided a simple index for how well a model generalizes: how much it can learn from how little data. We illustrated the usefulness of this index in a comparison of several current models over several formal languages. Beyond showing which current models generalize better than others, the benchmark also highlights which aspects of artificial neural networks work well for grammar induction, and what is still missing.

Among languages that were learned with perfect accuracy ($a^nb^n$, $a^nb^mc^{n+m}$, Dyck-1), MDLRNNs generalized best, but still failed on others ($a^nb^nc^n$, $a^nb^nc^nd^n$, and Dyck-2). Previous work has shown that this model's search procedure, an evolutionary algorithm, fails to find networks that are manually shown to have better MDL scores \cite{lan:2022a}.
We take this to show that the optimization procedure limits the model and prevents it from taking full advantage of the MDL objective.
The benefit of the MDL objective is nevertheless evident in the generalization performance for several languages.

MARNNs clearly benefit from their memory devices and reach good generalization scores, but testing for perfect accuracy ($\varepsilon=0$) reveals that their learning outcome is mostly approximate, and that they fail to maintain perfect accuracy for long stretches beyond their training data.
This could be the result of an inadequate objective function (cross-entropy), limitations of the search (backpropagation/gradient descent), or both. We do not currently have results that help decide this matter, but recent results for other architectures \cite{el-naggar:2023a} hint that the problem  lies at least in part in the objective function. 

\section{Acknowledgements}

This work was granted access to the HPC resources of IDRIS under the allocation 2023-AD011013783 made by GENCI.

\bibliography{bliss}

\appendix

\section{Appendix: Hyper-parameters}
\label{appendix:hyperparams}

\subsection{Training corpora}

All training sets were generated using the same random seed 100 and prior probability $p=0.3$. The datasets are available at \href{https://github.com/taucompling/bliss}{https://github.com/taucompling/bliss}. Following \citet{jacovi:2023}, the datasets are zipped and password-protected to prevent test data contamination of large language models through crawling.

Each of the LSTM and MARNN hyper-param configurations below was run 3 times using different random seeds (100, 101, 102). MDLRNNs were run once per configuration because of their longer running time.

\subsection{LSTM}

LSTMs were trained based on the following hyper-params grid: hidden state size (2/32/128), regularization technique (L1/L2/none), and the regularization constant in case regularization was applied ($\lambda$ = 1.0/0.1/0.01). Networks were trained using the Adam optimizer \cite{kingma:2017} with learning rate 0.001, $\beta_1 = 0.9$, and $\beta_2 = 0.999$. The networks were trained by feeding the full batch of training data for 1,000 epochs.

\subsection{MARNN}

MARNNs were trained by varying the architecture type (Softmax Stack-LSTM/Softmax Baby-NTM) and stack/memory size (50/100 for Stack-LSTM, 2050 for Baby-NTM). For Stack-LSTM, stack sizes were selected so they were always larger than the largest values seen during training: $n+m=22+24$ for $a^nb^mc^{n+m}$ and $n=24$ for all other languages. During testing the stack size was enlarged to 2050, beyond the maximum needed to reach scores $\bb{B}=$ 1 and 2. Baby-NTM memory was set to 2050 already during training because this model's memory size affects the weight dimensions and cannot be changed after training.

The rest of the hyper-parameters were set to the default values from \citet{suzgun:2019}. Stack-LSTM: hidden size 8; 1 layer; memory dimension 5; epochs 3/50; learning rate 0.01; Baby-NTM: hidden size 8; 1 layer; memory dimension 5; epochs 3/50; learning rate 0.01.

The original MARNN setup was modified here so that the network outputs represent probability distributions and not multi-label outputs.
This was done by replacing the sigmoid outputs with a softmax layer and the MSE loss with cross-entropy.

\subsection{MDLRNN}

MDLRNNs were trained using the evolutionary algorithm and the same hyper-params reported in \citet{lan:2022a}: population size 500; islands size 250; 25,000 generations; tournament size 2; early stop after 2 hours of no improvement; elite ratio 0.001; migration interval 1,000 generations/30 minutes.

\section{Appendix: PCFGs}

\label{appendix:pcfgs}

\subsection{$a^nb^n$}

\[
S \rightarrow \Biggl\{ 
	\begin{array}{lll}
		a S b &  & 1-p \\
		\varepsilon & & p
		
	\end{array}
\]

\subsection{$a^nb^mc^{n+m}$}

\[
\begin{array}{l}
S \rightarrow \Biggl \{ 
	\begin{array}{lll}
		a S c &  & 1-p \\
		X & & p
	\end{array}
\\
\\
X \rightarrow \Biggl \{ 
	\begin{array}{lll}
		b X c &  & 1-p \\
		\varepsilon & & p
	\end{array}
\end{array}
\]

\subsection{Dyck-1}

\[
S \rightarrow \Biggl \{ 
	\begin{array}{lll}
		(\ S\ )\ S &  & p \\
		\varepsilon & & 1-p
	\end{array}
\]

\subsection{Dyck-2}

\[
S \rightarrow \Biggl \{ 
	\begin{array}{lll}
		(\ S\ )\ S &  & p/2 \\
		\left[\ S\ \right]\ S &  & p/2 \\
		\varepsilon & & 1-p
	\end{array}
\]

\end{document}